\documentclass[11pt]{article}
\usepackage{graphicx}
\usepackage{mtsummit2019}
\usepackage{times}
\usepackage{url}
\usepackage{latexsym}
\usepackage[small,bf]{caption} 
\usepackage{comment}
\usepackage{array, booktabs}
\usepackage{amsmath,amssymb,bm}
\usepackage{float}
\usepackage{multirow}

\newcommand{\argmax}{\mathop{\rm argmax}\limits}

\title{
  Debiasing Word Embeddings Improves Multimodal Machine Translation
}

\author{
  Tosho Hirasawa\\
  Tokyo Metropolitan University\\
  {\tt hirasawa-tosho@ed.tmu.ac.jp}  \And
  Mamoru Komachi\\
  Tokyo Metropolitan University\\
  {\tt komachi@tmu.ac.jp}
}

\date{}

\begin{document}
\maketitle

\begin{abstract}
  In recent years, pretrained word embeddings have proved useful for multimodal neural machine translation (NMT) models
  to address the shortage of available datasets.
  However, the integration of pretrained word embeddings has not yet been explored extensively.
  Further, pretrained word embeddings in high dimensional spaces have been reported to suffer from the hubness problem.
  Although some debiasing techniques have been proposed to address this problem for other natural language processing tasks,
  they have seldom been studied for multimodal NMT models.
  In this study, we examine various kinds of word embeddings and introduce two debiasing techniques
  for three multimodal NMT models and two language pairs --- English--German translation and English--French translation.
  With our optimal settings, the overall performance of multimodal models was improved by up to 
  +1.62 BLEU and +1.14 METEOR for English--German translation and 
  +1.40 BLEU and +1.13 METEOR for English--French translation.
\end{abstract}

\section{Introduction}

In \textbf{multimodal machine translation}, a target sentence is translated from a source sentence 
together with related nonlinguistic information such as visual information.
Recently, neural machine translation (NMT) has superseded traditional statistical machine translation owing to the introduction of the attentive encoder-decoder model, in which machine translation is treated as a sequence-to-sequence learning problem and is trained to pay attention to the source sentence while decoding \cite{bahdanau2015jointly}.

Pretrained word embeddings are considered an important part of neural network models in many natural language processing (NLP) tasks.
In the context of NMT, pretrained word embeddings have proved useful in low-resource domains \cite{qi2018when},
in which FastText \cite{bojanowski2016enriching} embeddings are used to initialize the encoder and decoder of the NMT model.
They provided substantial overall performance improvement for low-resource language pairs.
Similarly, \newcite{hirasawa2019prediction} introduced a multimodal NMT model with embedding prediction 
that provided substantial performance improvement.

However, when word embeddings are used in the $k$-nearest neighbor ($k$NN) problem,
certain words appear frequently in the $k$-nearest neighbors for other words \cite{dinu2014improving,faruqui2016problems};
this is called the hubness problem in the general machine learning domain \cite{radovanovic2010hubs}.
This phenomenon harms the utility of pretrained word embeddings.
In the context of NMT, \newcite{rios2017proceedings} reported that NMT models produce less-accurate translations for less-frequent words, 
but they are not aware of the hubness problem in word embeddings.
Instead, they proposed annotating sense labels or lexical labels to address this problem.
However, it is known to be effective to debias word embeddings based on their local bias \cite{hara2015localized} or global bias \cite{mu2018allbutthetop} for word analogy tasks,
which does not require extra expensive annotations and references.

In this study, we explore the utility of well-established word embeddings and introduce debiasing techniques for multimodal NMT models.
The main contributions of this study are as follows:
\begin{enumerate}
  \item We show that GloVe word embeddings are useful for various multimodal NMT models 
        irrespective of the extent to which visual features are used in them.
  \item We introduce All-but-the-Top debiasing technique for pretrained word embeddings 
        to further improve multimodal NMT models.
\end{enumerate}

\section{Related Works}

With the recent development of multimodal parallel corpora such as Multi30K \cite{elliott2016multi30k}, 
many multimodal NMT models have been proposed.
Most of these models are divided into two categories: visual feature integration and multitask learning.
In both categories, visual features are extracted using image processing techniques.

\paragraph{Visual feature adaptation}

Visual features are extracted using image processing techniques and then integrated into a machine translation model in many ways.
These studies include incorporation with visual features in NMT models \cite{calixto2017doubly,zhou2018vagnmt}
and multitask learning models \cite{elliott2017imagination,zhou2018vagnmt},
as discussed later in Section \ref{mnmt}.

\paragraph{Data augmentation}

Owing to the lack of the available datasets, data augmentation is widely studied in multimodal NMT.
Compared to a parallel corpus without images \cite{gronroos2018memad} and a pseudo-parallel corpus \cite{helcl2018cuni},
few studies have used monolingual data.
\newcite{hirasawa2019prediction} proposed a multimodal NMT model with embedding prediction to fully use pretrained word embeddings.
However, the use of word embeddings has not been studied among various multimodal NMT models.
We examine three different word embeddings for three multimodal NMT models.

\section{Multimodal Neural Machine Translation\label{mnmt}}

In this study, we measure the effectiveness of pretrained word embeddings for 
doubly-attentive NMT \cite{calixto2017doubly}, 
IMAGINATION \cite{elliott2017imagination}, 
and visual attention grounding NMT \cite{zhou2018vagnmt};
these use visual feature integration, multitask learning, and mixed model, respectively.

First, in visual feature integration, visual features are incorporated into NMT models in different ways.
\newcite{calixto2017doubly} separately calculate textual and visual context vectors using an attention mechanism 
and then forward the concatenated context vector to output the probabilities of target words.
\newcite{caglayan2018liumcvc} use hidden states in the encoder to mask the local visual features and concatenate the textual context vector and the masked visual context vector to obtain the final context vector.

Second, in multitask learning, 
most multitask learning models use latent space learning as an auxiliary task.
Models share the encoder between the main translation task and the auxiliary task,
thereby improving the encoder.
\newcite{elliott2017imagination} proposed the IMAGINATION model
that learns to construct the corresponding visual feature from the hidden states of the textual encoder of a source sentence.

Third, visual feature integration and multitask learning are not mutually exclusive and can be used together.
\newcite{zhou2018vagnmt} compute the text representation from a source sentence while paying attention to each word based on the paired image.
This text representation is used in both the machine translation task and the shared space learning task.

All of these models tackle machine translation as a sequence-to-sequence learning problem
in which a neural model is trained to translate a source sentence of $N$--tokens $x = \{x_1, x_2, \cdots, x_N\}$ into the target sentence of $M$--tokens $y = \{y_1, y_2, \cdots, y_M\}$.

\subsection{Doubly-attentive NMT\label{calixto2017doubly}}

Doubly-attentive NMT \cite{calixto2017doubly} has a simple encoder and a modified decoder from \newcite{bahdanau2015jointly}
that uses two individual attention mechanisms to compute the textual context vector and the visual context vector.

\paragraph{Architecture}

The encoder is a bidirectional gated recurrent unit (GRU) \cite{cho2014gru}, 
in which a forward GRU encodes source sentence $x$ in the normal order to generate a sequence of forward hidden states $\overrightarrow{\bm{h}} = \{\overrightarrow{\bm{h}_1}, \overrightarrow{\bm{h}_2}, \cdots, \overrightarrow{\bm{h}_N}\}$
and a backward GRU encodes this source sentence in the reversed order to generate a sequence of backward hidden states $\overleftarrow{\bm{h}} = \{\overleftarrow{\bm{h}_1}, \overleftarrow{\bm{h}_2}, \cdots, \overleftarrow{\bm{h}_N}\}$.
The final hidden states $\bm{h}$ for each position $i$ are given as a concatenation of each forward hidden state and each backward hidden state.

\begin{eqnarray}
  \overrightarrow{\bm{h}_i} &=& \overrightarrow{\mathrm{GRU}}(
    \overrightarrow{\bm{h}_{i-1}}, \bm{e}_{enc}(x_i)) \\
  \overleftarrow{\bm{h}_i} &=& \overleftarrow{\mathrm{GRU}}(
    \overleftarrow{\bm{h}_{i+1}}, \bm{e}_{enc}(x_i)) \\
  \bm{h}_i &=& [\overrightarrow{\bm{h}_i}; \overleftarrow{\bm{h}_i}]
\end{eqnarray}
where $i\in[1, N]$ denotes each position in a source sentence; $\overrightarrow{\mathrm{GRU}}$ and $\overleftarrow{\mathrm{GRU}}$ are the forward and backward GRU, respectively;
and $\bm{e}_{enc}(x_i)$ is the embedding representation for a word $x_i$.

While decoding, the model first computes a hidden state proposal $\bm{s}_j$ for each time step $j\in[1, M]$.
\begin{eqnarray}
  \bm{s}_j = \mathrm{GRU}(\hat{\bm{s}}_{j-1}, \bm{e}_{dec}(\hat{y}_{j-1}))
\end{eqnarray}
where $\hat{\bm{s}}_{j-1}$ is the previous hidden state
and $\bm{e}_{dec}(\hat{y}_{j-1})$ is the embedding for the previous output word $\hat{y}_{j-1}$.

The textual context vector and the visual context vector are computed using two independent attention mechanisms.
In each time step $j$ while decoding, a feed-forward layer is used to calculate a normalized soft alignment $\alpha_{j,i}$ with each source hidden state $\bm{h}_i$,
and the textual context vector $\bm{c}_j^{t}$ is computed as a weighted sum of source hidden states.
\begin{eqnarray}
  z_{j,i}^{t} &=& \bm{v}_{t} \mathrm{tanh}(
    \bm{U}_{\alpha}^{t}\bm{s}_j + \bm{W}_{\alpha}^{t} \bm{h}_i
  ) \\
  \alpha_{j,i}^{t} &=& \frac{\mathrm{exp}(z_{j,i}^{t})}{\sum_{k=1}^N{\mathrm{exp}(z_{j,k}^{t})}} \\
  \bm{c}_j^{t} &=& \sum_{i=1}^N{\alpha_{j,i}^{t}\bm{h}_i}
\end{eqnarray}
where $\bm{v}_{t}$, $\bm{U}_{\alpha}^{t}$ and $\bm{W}_{\alpha}^{t}$ are model parameters.

The visual context vector $\bm{c}_j^{v}$ is also computed from the spatial visual features $\bm{v}_i$ of the paired image 
in the same manner as the textual context vector along with the gating scalar mechanism, 
in which a scalar variable is computed based on the previous hidden state to decide how much attention should be paid to the entire visual features.
\begin{eqnarray}
  z_{j,i}^{v} &=& \bm{v}_{v} \mathrm{tanh}(
    \bm{U}_{\alpha}^{v}\bm{s}_j + \bm{W}_{\alpha}^{v} \bm{v}_i
  ) \\
  \alpha_{j,i}^{v} &=& \frac{\mathrm{exp}(z_{j,i}^{v})}{\sum_{k=1}^N{\mathrm{exp}(z_{j,k}^{v})}} \\
  \beta_j &=& \mathrm{\sigma}(\bm{W}_s \bm{\hat{s}}_{j-1} + \bm{b}_s) \\
  \bm{c}_j^{v} &=& \beta_j \sum_{i=1}^N{\alpha_{j,i}^{v}\bm{v}_i}
\end{eqnarray}
where $\bm{v}_{v}$, $\bm{U}_{\alpha}^{v}$, $\bm{W}_{\alpha}^{v}$, $\bm{W}_s$, and $\bm{b}_s$ are model parameters.
$\mathrm{\sigma}$ is the gating scalar function learnt while training;
it projects a vector to a scalar value and activates with a sigmoid function.

The final hidden state $\bm{\hat{s}}_j$ is computed using the hidden state proposal $\bm{s}_j$, 
textual context $\bm{c}_j^{t}$, and visual context $\bm{c}_j^{v}$.
\begin{alignat}{1}
  &\bm{z}_j = \bm{\mathrm{\sigma}}_z(
    \bm{W}_z^{t} \bm{c}_j^{t} + \bm{W}_z^{v} \bm{c}_j^{v} + \bm{W}_z \bm{\hat{s}}_j
  ) \\
  &\bm{r}_j = \bm{\mathrm{\sigma}}_r(
    \bm{W}_r^{t} \bm{c}_j^{t} + \bm{W}_r^{v} \bm{c}_j^{v} + \bm{W}_r \bm{\hat{s}}_j
  ) \\
  &\bm{s'}_j = \mathrm{tanh}(
    \bm{W}_z^{t} \bm{c}_j^{t} + \bm{W}_z^{v} \bm{c}_j^{v} + 
    \bm{r}_j \odot (\bm{U} \bm{\hat{s}}_j)
  ) \\
  &\bm{\hat{s}}_j = (1 - \bm{z}_j) \odot \bm{s'}_j + \bm{z}_j \odot \bm{s}_j
\end{alignat}
where $\bm{\mathrm{\sigma}}_z$ and $\bm{\mathrm{\sigma}}_r$ are feed-forward layers with sigmoid activation,
and $\bm{W}_z^{t}$, $\bm{W}_z^{v}$, $\bm{W}_z$, $\bm{W}_r^{t}$, $\bm{W}_r^{v}$, $\bm{W}_r$, $\bm{W}_z^{t}$, $\bm{W}_z^{v}$, and $\bm{U}$ are model parameters.

The system output at timestep $j$ is obtained using the current hidden state, previous word embedding, textual context, and visual context.
\begin{equation}
  \begin{split}
    p(w|\hat{y}_{<j}) &= \mathrm{softmax}(
      \mathrm{tanh}(
        \bm{L}^s \bm{\hat{s}}_j \\ 
        &+ \bm{L}^w \bm{e}_{dec}(\hat{y}_{j-1}) + \bm{L}^t \bm{c}_j^{t} + \bm{L}^i \bm{c}_j^{v}
      )
    )      
  \end{split}
\end{equation}
\begin{equation}
  \hat{y}_j = \argmax_{w\in\mathcal{V}} \{ p(w|\hat{y}_{<j}) \}
\end{equation}
where $\bm{L}^s$, $\bm{L}^w$, $\bm{L}^t$ and $\bm{L}^i$ are model parameters.

\paragraph{Loss function}

We use the negative log likelihood of the probabilities to generate reference tokens as the loss function $J$ for this model.
\begin{equation}\label{nll_loss}
  J = - \sum_{j=1}^{M} \mathrm{log}( p(y_{j}|\hat{y}_{<j}))
\end{equation}

\subsection{IMAGINATION\label{imagination}}

IMAGINATION \cite{elliott2017imagination} is a multitask learning model 
that jointly learns machine translation and visual latent space models.
It trains an NMT model for a machine translation task and a latent space learning model for an auxiliary task, 
in which a source sentence and the paired image are mapped closely in the latent space.
The models for each task share the same encoder in a multitask scenario.

\paragraph{Architecture}

The encoder is the same as that in the doubly-attentive NMT model described in Section \ref{calixto2017doubly}.
The decoder in the NMT model is the same as that proposed by \newcite{bahdanau2015jointly};
it first computes the hidden state proposal $\bm{s}_j$, then estimates context vector $\bm{c}_j$ over source hidden states, and finally outputs the predicted word $y_j$ for each time step $j\in[1,M]$.
\begin{eqnarray}
  \bm{s}_j &=& \mathrm{GRU}(\hat{\bm{s}}_{j-1}, \bm{e}_{dec}(\hat{y}_{j-1})) \\
  z_{j,i} &=& \bm{v}_a \mathrm{tanh}(
    \bm{W}_a\bm{s}_j + \bm{U}_a \bm{h}_i
  ) \\
  \alpha_{j,i} &=& \frac{
    \mathrm{exp}(z_{j,i})
  }{
    \sum_{k=1}^N{\mathrm{exp}(z_{j,k})}
  } \\
  \bm{c}_j &=& \sum_{i=1}^{N} \alpha_{j,i} \bm{h}_i
\end{eqnarray}
\begin{equation}
  p(w|\hat{y}_{<j}) = \mathrm{softmax}(
    \mathrm{tanh}(
        \bm{s}_j + 
        \bm{e}_{dec}(\hat{y}_{j-1}) + \bm{c}_j
    )
  )
\end{equation}
\begin{equation}
  \begin{split}
    \hat{y}_j = \argmax_{w\in\mathcal{V}} \{ p(w|\hat{y}_{<j}) \}
  \end{split}
\end{equation}
where $\bm{W}_a$, $\bm{U}_a$ and $\bm{v}_a$ are model parameters.

The latent space learning model calculates the average vector over the hidden states $\bm{h_i}$ in the encoder and maps it to the final vector $\bm{\hat{v}}$ in the latent space.
\begin{equation}
    \bm{\hat{v}} = \tanh(\bm{W_\mathrm{v}} \cdot \frac{1}{N} \sum_i^N \bm{h_i} )
\end{equation}
where $\bm{W_\mathrm{v}}$ is a model parameter.

\paragraph{Loss function}

The loss function for IMAGINATION is the linear interpolation of loss functions of each task.
\begin{equation} \label{mtl_loss}
  J = \lambda J_\mathrm{T}(\theta,\phi_\mathrm{T}) + (1 - \lambda)J_\mathrm{V}(\theta,\phi_\mathrm{V})    
\end{equation}
where $\theta$ is the parameter of the shared encoder; $\phi_\mathrm{T}$ and $\phi_\mathrm{V}$ are parameters of the machine translation model and latent space model, respectively; and
$\lambda$ is the interpolation coefficient\footnote{We use $\lambda=0.5$ in our experiment.}.

We use the loss function defined in Eq. \ref{nll_loss} for the NMT model $J_\mathrm{T}(\theta, \phi_\mathrm{T})$.
\begin{equation}\label{nll_loss_mt}
  J_\mathrm{T}(\theta, \phi_\mathrm{T}) = - \sum_{j=1}^{M} \mathrm{log}( p(y_{j}|\hat{y}_{<j}))
\end{equation}

The max margin loss is used as the loss function for latent space learning;
it makes corresponding latent vectors of a source sentence and the paired image closer.
\begin{equation}
  J_\mathrm{V}(\theta,\phi_\mathrm{V}) = \sum_{\bm{v'} \neq \bm{v}} \max \{ 0, 
    \alpha - d(\bm{\hat{v}}, \bm{v}) + d(\bm{\hat{v}}, \bm{v'}) \}
\end{equation}
where $\bm{v}$ is the latent vector of the paired image;
$\bm{v'}$, the image vector for other examples;
$d$, the cosine similarity function that is used to calculate the word similarity;
and $\alpha$, the margin that adjusts the sparseness of each vector in the latent space\footnote{We use $\alpha=0.1$ in our experiment.}.

\subsection{Visual Attention Grounding NMT}

Visual Attention Grounding NMT (VAG-NMT) \cite{zhou2018vagnmt} uses a combination of the visual feature integration model and the multitask learning model, which also uses latent space learning as the auxiliary task.

\paragraph{Architecture}

The shared encoder of this model is an extension of \newcite{bahdanau2015jointly},
in which the model computes the sentence representation $\bm{t}$ by paying attention to the hidden states $\bm{h}_i$ based on the visual feature $\bm{v}$.
\begin{eqnarray}
  z_i &=& \mathrm{tanh}(\bm{W}_v \bm{v}) \cdot \mathrm{tanh}(\bm{W}_h \bm{h}_i) \\
  \beta_i &=& \frac{
    \mathrm{exp}(z_i)
  }{
    \sum_{k=1}^{N}\mathrm{exp}(z_k)
  } \\
  \bm{t} &=& \sum_{i=1}^{N} \beta_i \bm{h}_i
\end{eqnarray}
where $\bm{W}_v$ and $\bm{W}_h$ are model parameters.

The decoder of the NMT model is the same as that used in IMAGINATION (Section \ref{imagination}) 
with a slight modification for initializing the hidden state with the sentence representation $\bm{t}$.
\begin{eqnarray}
  \bm{s}_0 = \mathrm{tanh}(\bm{W}_{init}(
    \rho \bm{t} + (1 - \rho) \frac{1}{N}\sum_{i}^{N} \bm{h}_i
  ))
\end{eqnarray}
where $\bm{W}_{init}$ is a model parameter; 
and $\rho$, a hyperparameter to determine the ratio of text representation in the decoder initial state \footnote{We use $\rho = 0.5$ in our experiment.}.

In latent space learning, both the sentence representation $\bm{t}$ and the visual representation $\bm{v}$ are projected to the latent space and made closer in the space during training.
\begin{eqnarray}
  \bm{t}_{emb} &=& \mathrm{tanh}(\bm{W}_t \bm{t} + \bm{b}_t) \\
  \bm{v}_{emb} &=& \mathrm{tanh}(\bm{W}_v \bm{v} + \bm{b}_v)
\end{eqnarray}
where $\bm{W}_t$, $\bm{b}_t$, $\bm{W}_v$, and $\bm{b}_v$ are model parameters.

\paragraph{Loss function}

The loss function for VAG-NET is given as described in Eq.\ref{mtl_loss}, 
and we use the loss function defined in Eq.\ref{nll_loss_mt} for $J_\mathrm{T}(\theta, \phi_\mathrm{T})$.

The max margin loss with negative sampling is used as the loss function for latent space learning.
\begin{equation}
  \begin{split}
    &J_\mathrm{V}(\theta,\phi_\mathrm{V}) \\
    &= \sum_{p} \sum_{k} \max \{0, \gamma - d(\bm{v}_p, \bm{t}_p) + d(\bm{v}_p, \bm{t}_{k \neq p}) \} \\
    &+ \sum_{k} \sum_{p} \max \{0, \gamma - d(\bm{t}_k, \bm{v}_k) + d(\bm{t}_p, \bm{v}_{k \neq p}) \}
  \end{split}
\end{equation}
where $d$ is a cosine similarity function; 
$k$ and $p$ is the index for sentences and images, respectively;
$\bm{t}_{k \neq p}$, the negative samples for which all examples in the same batch with the target example are selected;
and $\gamma$, the margin that adjusts the sparseness of each item in the latent space\footnote{We use $\gamma = 0.1$ in our experiment.}.

\section{Word Embedding}

In this study, we compare three different word embeddings: 
word2vec \cite{mikolov2013distributed}, GloVe \cite{pennington2014glove}, and FastText \cite{bojanowski2016enriching}.
Section \ref{embedding_setting} describes the configurations to build each embedding.

When we use word embeddings of high dimension in the $k$NN problem 
in which the similarity of two words is computed using a distance function,
certain words frequently appear in the $k$-nearest neighbors of other words \cite{dinu2014improving,faruqui2016problems};
this is called the hubness problem in the general machine learning domain \cite{radovanovic2010hubs}.
This phenomenon harms the utility of pretrained word embeddings.
In the context of NMT, \newcite{rios2017proceedings} report that less-frequent words are translated with low-accuracy;
that may be influenced by the hubness problem.

To address this problem, localized centering \cite{hara2015localized} and All-but-the-Top \cite{mu2018allbutthetop} have been proposed in  NLP literature,
in which pretrained word embeddings are debiased using the local bias of each word or the global bias of the entire vocabulary.
In this study, both debiasing techniques are tested for all embedding types.

\paragraph{Localized centering}

Localized centering shifts each word based on its local bias.
The local centroid for each word $x$ is computed and subtracted from the original word $x$ to obtain the new embedding $\hat{x}$.
\begin{eqnarray}
  c_k(x) &=& \frac{1}{k} \sum_{x' \in k\mathrm{NN}(x)} x' \\
  \hat{x} &=& x - c_k(x)
\end{eqnarray}
where $k$ is a hyperparameter called local segment size \footnote{We use $k=10$ in our experiment.};
$k\mathrm{NN}(x)$ returns the $k$--nearest neighbors of the word $x$.

\paragraph{All-but-the-Top}

All-but-the-Top uses the global bias of the entire vocabulary to shift the embedding of each word.
The algorithm of All-but-the-Top has three steps:
subtract the centroid of all words from each word $x$,
compute the PCA components for the centered space,
and subtract the top $n$ PCA components from each centered word to obtain the final word $\hat{x}$.
\begin{eqnarray}
  x' &=& x - \frac{1}{|\mathcal{V}|} \sum_{w \in \mathcal{V}} w \\
  u_1, u_2, \cdots, u_D &=& \mathrm{PCA}(x' \in \mathcal{V}) \\
  \hat{x} &=& x' - \sum_{i=1}^{D} (u_i^\mathsf{T} x')u_i
\end{eqnarray}
where $D$ is a hyperparameter that is used to determine how many principal components of pretrained word embeddings are ignored\footnote{We use $D=3$ in our experiment.}.

\section{Experiment}

\subsection{Word Embeddings}\label{embedding_setting}

\paragraph{Training corpus}

As publicly available pretrained word embeddings use different training corpora,
we created a monolingual corpus from Wikipedia for a fair comparison.
We downloaded the January 20, 2019, version of Wikidump for English, German, and French\footnote{https://dumps.wikimedia.org/} and extracted article pages.
All extracted sentences are preprocessed by lower-casing, tokenizing, and normalizing the punctuation using the Moses script \footnote{
  We applied preprocessing using \texttt{task1-tokenize.sh} from https://github.com/multi30k/dataset.
}.
Table \ref{wiki_stats} shows the statistics of the preprocessed Wikipedia corpus for each language.

\begin{table}[t]
  \begin{center}
      \begin{tabular}{l|rrr} \toprule
          Language  & Lines & Types & Tokens \\ \midrule[0.08em]
          English   & 96M   & 10M   & 2,347M \\
          German    & 35M   & 11M   & 829M   \\
          French    & 39M   & 4M    & 703M   \\ \bottomrule
      \end{tabular}{}
  \end{center}
  \caption{
      Statistics of Wikipedia corpus for each language.
  }
  \label{wiki_stats}
\end{table}

\paragraph{Training settings}

All embeddings trained on Wikipedia have a dimension of 300.
The specific options set for training are as follows;
default values were used for other options.

We trained the word2vec model\footnote{We train using https://github.com/tmikolov/word2vec.} using the CBOW algorithm with window size of 10, negative sampling of 10, and minimum count of 10;
the GloVe model\footnote{We train using https://github.com/stanfordnlp/GloVe.} with windows size of 10 and minimum count of 10;
and the FastText model\footnote{We train using https://github.com/facebookresearch/fastText.} using the CBOW algorithm with word n-gram of 5, window size of 5, and negative sampling of 10.

\paragraph{Unknown words}

There are two types of unknown words:
words that are a part of pretrained word embeddings but are not included in a vocabulary (Out-Of-Vocabulary (OOV) words)
and words that are a part of a vocabulary but are not included in pretrained word embeddings (OOV words for embeddings).
OOV words for embeddings only exist when using word-level embeddings (word2vec and glove);
the embeddings of such words in FastText are calculated as the mean embedding of character n-grams consisting of the word.

The embeddings for both types of OOV words are calculated as the average embedding over words 
that are a part of pretrained word embeddings but are not included in the vocabularies,
and they are updated individually during training.

\subsection{Dataset}

We train, validate, and test all multimodal NMT models using the Multi30K \cite{elliott2016multi30k} dataset.
English is selected as the source language, and German/French are selected as target languages.
All sentences in all languages are preprocessed by lower-casing, tokenizing, and normalizing the punctuation.
 
We run experiments without byte pair encoding (BPE) \cite{sennrich2016bpe} for all models
as BPE breaks a word into subwords, resulting in an increase in OOV words for word2vec and GloVe embeddings.
In addition, we also run experiments using BPE with 10k merge operations to show the utility of pretrained word embeddings.
The BPE subwords are shared for source and target languages and learnt from training dataset\footnote{
  We use https://github.com/rsennrich/subword-nmt to train and apply BPE.
}.
Table \ref{vocab_stats} shows the statistics of vocabularies in the Multi30K training data.

Visual features are extracted using pretrained ResNet-50 \cite{he2016deep}.
We encode all images in the Multi30K dataset using ResNet-50 and
pick out the hidden state in the res4f layer of 1024D for the doubly-attentive model,
and that in the pool5 layer of 2048D for IMAGINATION and VAG-NET, respectively.

\begin{table}[t]
  \begin{center}
      \begin{tabular}{l|rrrr} 
        \toprule
                  &        &         & \multicolumn{2}{c}{OOV} \\
        Language  & Types  & Tokens  & Vocab & Embed   \\ 
        \midrule[0.08em]
        English   & 10,210 & 377,534 & 10M   & 129   \\
        German    & 18,722 & 360,706 & 11M   & 1,841 \\
        French    & 11,219 & 409,845 & 4M    & 89    \\  
        \midrule[0.08em]
        \multicolumn{5}{c}{with BPE} \\
        \midrule[0.08em]
        English       & 5,199 & 397,793 & N/A & N/A \\
        $\to$ German  & 7,062 & 400,507 & N/A & N/A \\
        English       & 5,830 & 394,353 & N/A & N/A \\
        $\to$ French  & 6,572 & 428,762 & N/A & N/A \\
        \bottomrule
      \end{tabular}{}
  \end{center}
  \caption{
      Statistics of vocabularies without BPE (upper) and with BPE (lower) in Multi30K training data.
      ``Vocab'' denotes the number of OOV words for the vocabulary.
      ``Embed'' denotes the number of OOV words for embeddings.
      ``English $\to$ German'' shows statistics of the shared subwords for English--German translation,
      and ``English $\to$ French'' for English--French translation.
  }
  \label{vocab_stats}
\end{table}

\subsection{Model}

All models are implemented using nmtpytorch toolkit v3.0.0\footnote{https://github.com/toshohirasawa/mmt-emb-init} \cite{caglayan2017nmtpy}.

The encoder for each model has one layer with 256 hidden dimensions, 
and therefore the bidirectional GRU has 512 dimensions.
We set the latent space vector size for IMAGINATION to 2048 and
the dimension of the shared visual-text space for VAG-NET to 512.
The input word embedding size and output vector size are 300 each.

We use the Adam optimizer with learning rate of 0.0004.
The gradient norm is clipped to 1.0. 
The dropout rate is 0.3.

BLEU \cite{papineni2002bleu} and METEOR \cite{denkowski2014meteor} are used as performance metrics.
As in \cite{qi2018when}, we also evaluated the models using the F-score of each word.
The F-score is calculated as the harmonic mean of the precision 
(the fraction of produced sentences containing a word that is in the references sentences)
and the recall (the fraction of reference sentences containing a word that is in the model outputs).
We ran the experiment three times with different random seeds and obtained the mean for each model.

\subsection{Results}

\begin{table*}
  \begin{center}
      \begin{tabular}{ll|cc|cc|cc} \toprule
        \multicolumn{8}{c}{\textbf{English $\to$ German}} \\
        \midrule[0.08em]
                          & debiasing&\multicolumn{2}{c}{None} & \multicolumn{2}{c}{Localized Centering} & \multicolumn{2}{c}{All-but-the-Top} \\
        Model             & embedding & BLEU  & METEOR & BLEU  & METEOR & BLEU  & METEOR  \\
        \midrule[0.08em]
        NMT               & random    & 34.57 & 54.50 & \multicolumn{4}{c}{} \\
        Doubly-attentive  & random    & 33.50 & 52.75 & \multicolumn{4}{c}{} \\
        IMAGINATION       & random    & 34.97 & 54.21 & \multicolumn{4}{c}{} \\
        VAG-NET           & random    & 35.55	& 54.87 & \multicolumn{4}{c}{} \\
        \midrule[0.08em]
        NMT               & word2vec  & 34.23 & 52.83 & 34.14 & 53.09  & 33.88 & 52.66 \\
                          & GloVe     & 35.49	& 55.14 & 35.33 & 54.89  &	\textbf{35.98}	& \textbf{55.15} \\
                          & FastText  & 33.63	& 52.48	& 33.42 & 52.34  &	33.91	& 52.65 \\
        \midrule
        Doubly-attentive  & word2vec  & 32.05	& 50.85	& 32.07 & 51.23  & 32.73	& 51.04 \\
                          & GloVe     & 34.06	& 53.74 & 33.37 & 52.98  &	\textbf{34.77}	& \textbf{53.86} \\
                          & FastText  & 31.14	& 49.29	& 31.04 & 50.33  & 30.86	& 50.13 \\
        \midrule
        IMAGINATION       & word2vec  & 33.97	& 52.59 & 33.43 & 52.32  & 34.35	& 52.79 \\
                          & GloVe     & 35.74	& 55.00	& 35.92 & 55.15  &	\textbf{36.59}	& \textbf{55.35} \\
                          & FastText  & 34.21	& 52.53 & 33.69 & 52.22  & 33.83	& 52.31 \\
        \midrule
        VAG-NET           & word2vec  & 34.32	& 53.01	& 34.10	& 53.40  & 33.91 & 52.70 \\
                          & GloVe     & 36.01	& \textbf{55.31} & 35.56	& 54.61  & \textbf{36.36} & 55.17 \\
                          & FastText  & 34.12	& 52.56	& 33.92 & 52.75  & 33.82 & 52.38 \\ 
        \bottomrule
      \end{tabular}{}
  \end{center}
  \caption{
      Results obtained using Multi30K test2016 dataset for English--German translation.
      ``NMT'' shows the results of \newcite{bahdanau2015jointly}.
      When the debiasing is ``None,'' we show the results obtained with raw pretrained word embeddings or random values.
  }
  \label{results_ende}
\end{table*}

\begin{table*}[tp]
  \begin{center}
      \begin{tabular}{ll|cc|cc|cc} \toprule
        \multicolumn{8}{c}{\textbf{English $\to$ French}} \\
        \midrule[0.08em]
                          & debiasing&\multicolumn{2}{c}{None} & \multicolumn{2}{c}{Localized Centering} & \multicolumn{2}{c}{All-but-the-Top} \\
        Model             & embedding & BLEU  & METEOR & BLEU  & METEOR & BLEU  & METEOR  \\
        \midrule[0.08em]
        NMT               & random    & 57.15	& 72.47 & \multicolumn{4}{c}{} \\
        Doubly-attentive  & random    & 54.85	& 71.06 & \multicolumn{4}{c}{} \\
        IMAGINATION       & random    & 57.38	& 72.57 & \multicolumn{4}{c}{} \\
        VAG-NET           & random    & 57.78	& 73.21 & \multicolumn{4}{c}{} \\
        \midrule[0.08em]
        NMT               & word2vec  & 55.65	& 70.79	& 55.82	& 70.90	& 56.20	& 71.20 \\
                          & GloVe     & 58.14	& \textbf{73.67}	& 57.76	& 73.00	& \textbf{58.24}	& 73.40 \\
                          & FastText  & 55.13	& 70.18	& 55.24	& 70.56	& 55.42	& 70.60 \\
        \midrule
        Doubly-attentive  & word2vec  & 52.32	& 68.06	& 53.30	& 68.98	& 52.95	& 68.68 \\
                          & GloVe     & \textbf{56.25} & \textbf{72.19}	& 54.58	& 71.23	& 56.12	& 71.91 \\
                          & FastText  & 50.46	& 66.35	& 51.02	& 67.20	& 51.22	& 67.09 \\
        \midrule
        IMAGINATION       & word2vec  & 55.94	& 70.91	& 55.63	& 70.73	& 55.96	& 70.93 \\ 
                          & GloVe     & 57.89	& 73.09	& 57.65	& 73.16	& \textbf{58.10}	& \textbf{73.26} \\ 
                          & FastText  & 55.12	& 70.17	& 55.52	& 70.77	& 55.52	& 70.42 \\ 
        \midrule
        VAG-NET           & word2vec  & 56.23	& 71.14	& 55.79	& 70.82	& 56.33	& 71.34 \\
                          & GloVe     & \textbf{58.45}	& \textbf{73.59}	& 57.31	& 73.16	& 57.94	& 73.40 \\
                          & FastText  & 55.25	& 70.45	& 55.33	& 70.51	& 55.49	& 70.63 \\ 
        \bottomrule
      \end{tabular}{}
  \end{center}
  \caption{
    Results obtained using Multi30K test2016 dataset for English--French translation.
    ``NMT'' shows the results of \newcite{bahdanau2015jointly}.
    When the debiasing is ``None,'' we show the results obtained with raw pretrained word embeddings or random values.
  }
  \label{results_enfr}
\end{table*}

\begin{figure*}[t]
  \begin{minipage}{0.5\hsize}
    \begin{center}
      \includegraphics[width=8cm]{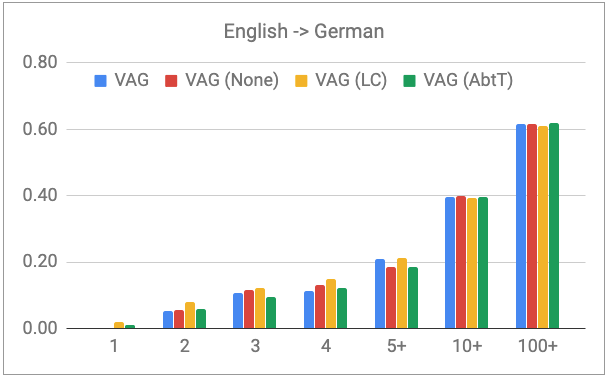}      
    \end{center}
  \end{minipage}
  \begin{minipage}{0.5\hsize}
    \begin{center}
      \includegraphics[width=8cm]{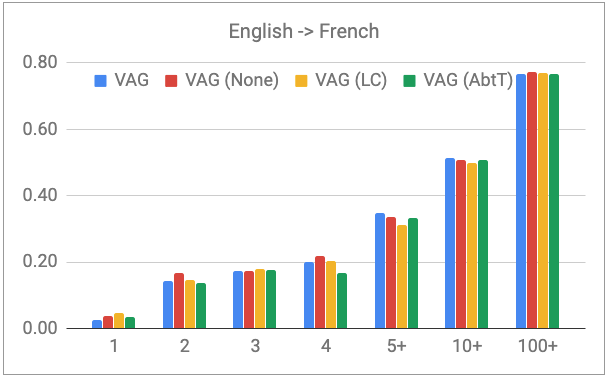}
    \end{center}
  \end{minipage}
  \caption{
      F-score of word prediction per frequency breakdown in training corpus.
      The model without brackets is initialized with random values:
      ``(None),'' GloVe without debiasing;
      ``(LC),'' GloVe with localized centering; and
      ``(AbtT),'' GloVe with All-but-the-Top.
  }
  \label{fscore}
\end{figure*}

\renewcommand{\arraystretch}{1.4}
\begin{table*}[t]
  \begin{center}
      \begin{tabular}{l|l|p{7cm}} \toprule
          \multirow{4}{*}{
            \begin{minipage}{5cm}
              \centering
              \includegraphics[width=5cm]{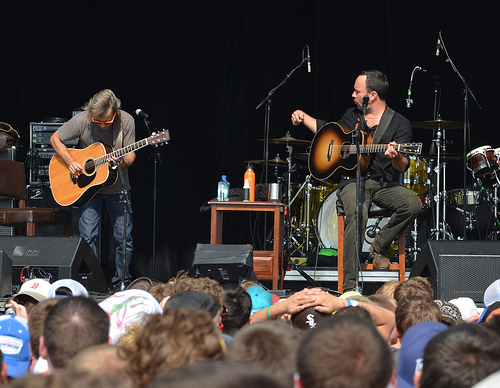}
            \end{minipage}
          } &
          Source &
          two men playing guitar in front of a large audience . \\
          &
          Reference &
          zwei männer spielen gitarre vor einem großen publikum . \\
          &
          VAG &
          zwei männer spielen vor einem großen publikum gitarre . \\
          &
          VAG (GloVe) &
          zwei männer spielen gitarre vor einem großen publikum . \\ \bottomrule
      \end{tabular}{}
  \end{center}
  \caption{
    Examples of English--German translations obtained using test dataset.
    ``(GloVe)'' denotes the model with the optimal settings for GloVe.
  }
  \label{model_examples}
\end{table*}
\renewcommand{\arraystretch}{1}

Table \ref{results_ende} shows the overall performance of the randomly initialized models 
and the models initialized with pretrained word embeddings for English--German translation.
Though GloVe embeddings show considerable improvement for both in text-only NMT and all types of multimodal NMT, 
word2vec and FastText embeddings greatly reduce model performance even with some debiasing.
With GloVe embeddings, All-but-the-Top debiasing results in further improvement. 
In particular, IMAGINATION is improved with GloVe embedding initialization (+0.77 BLEU and +0.79 METEOR) 
and showed further improvement with All-but-the-Top debiasing (+1.62 BLEU and +1.14 METEOR).

Table \ref{results_enfr} shows that 
the combination of GloVe embedding and All-but-the-Top debiasing greatly improves the overall performance of each model for English--French translation.
The model with GloVe and All-but-the-Top surpasses the randomly initialized model 
by +1.09 BLEU and +0.93 METEOR in the text-only NMT model,
by +1.27 BLEU and +0.85 METEOR in the doubly-attentive model,
by +0.72 BLEU and +0.69 METEOR in the IMAGINATION model,
by +0.16 BLEU and +0.19 METEOR in the VAG-NET model, respectively.

\begin{table}[t]
  \begin{center}
    \begin{tabular}{c}
      \begin{minipage}[c]{0.5\textwidth}
        \centering
        \begin{tabular}{ll|cc}
          \toprule
          \multicolumn{4}{c}{\textbf{English $\to$ German}} \\
          \midrule[0.08em]
          BPE   & Init    & BLEU  & METEOR \\
          \midrule[0.08em]
          No    & random  & 35.55 & 54.87 \\
          No    & GloVe   & \textbf{36.36} & \textbf{55.17} \\
          Yes   & random  & 35.46	& 55.30 \\
          \bottomrule
        \end{tabular}{}
      \end{minipage} \\
      \\
      \begin{minipage}[c]{0.5\textwidth}
        \centering
        \begin{tabular}{ll|cc}
          \toprule
          \multicolumn{4}{c}{\textbf{English $\to$ French}} \\
          \midrule[0.08em]
          BPE & Init   & BLEU  & METEOR \\
          \midrule[0.08em]
          No  & random & 57.78 & 73.21 \\
          No  & GloVe  & \textbf{58.45} & \textbf{73.59} \\
          Yes & random & 56.63 & 72.38 \\
          \bottomrule
        \end{tabular}{}
      \end{minipage}
    \end{tabular}{}
  \end{center}
  \caption{
    Results of VAG-NET with various settings obtained using Multi30K test2016 dataset 
    for English--German translation (upper) and English--French translation (lower).
    ``BPE'' denotes whether a model uses BPE.
    ``Init'' denotes the initialization strategy:
    ``random,'' a model initialized using random values and
    ``GloVe,'' a model initialized using GloVe embeddings with All-but-the-Top debiasing (English--German) or without debiasing (English--French).
  }
  \label{bpe_comparison}
\end{table}

\section{Discussion}

\paragraph{Word embedding}

In our study, GloVe performs the best among three word embeddings, while word2vec and FastText do not help multimodal NMT models;
the degradation of word2vec is attributed to the cohesiveness of word embeddings and that of FastText the shortage of training data, respectively.

The word embeddings in word2vec are reported to be cohesively clustered and not evenly distributed,
while those in GloVe are well distributed \cite{mimno2017geometry}.
This makes it harder to train the model with word2vec rather than the model initialized using random values,
as the model with word2vec is required to learn all the word representations from almost the same value i.e. the mean vector of entire embeddings.

FastText requires more training data than GloVe does,
as it learns not only embeddings for words but also those of their subwords.
Our pretrained word embeddings are trained using only Wikipedia and do not use Common Crawl;
it contains at least 50 times tokens and three times words than Wikipedia does, 
and is used together with Wikipedia to construct FastText embeddings that improve NMT models \cite{qi2018when}.

\paragraph{Debiasing}

All-but-the-Top improves most of models for both English--German translation and English--French translation;
this may prove the idea suggested in \newcite{mu2018allbutthetop},
in which neural network models may not be able to learn the debiasing technique by themselves.

In contrast, models using localized centering only show a comparable performance with models not using debiasing.
It is because that the debiased vector has small norm and thus the additional training may break the relation of debiased vectors,
as localized centering subtracts the local centroid of a word that is quite similar with the word.
This observation is contrary with \newcite{hara2015localized},
in which debiased word embeddings are used without the additional training.

\paragraph{Languages}

We found that pretrained word embeddings are more useful for English--German translation than for English--French translation.
The best models with GloVe embedding surpasses the randomly initialized model by +1.28 BLEU in average for English--German translation,
but by only +0.97 BLEU for English--French translation with the optimal settings.
This is because the German decoder has more unique words (18,722 for German and 11,219 for French, as listed in Table \ref{vocab_stats}) in the original training dataset, resulting in less in-vocabulary words after restricting the vocabulary
and making it difficult for the German decoder learn embeddings from scratch.

\paragraph{BPE}
BPE is an alternative approach to improve translation quality.
Therefore, we compared the VAG-NET model with GloVe embeddings and the VAG-NET model with BPE to validate which approach would contribute more to the overall performance (Table \ref{bpe_comparison}).
Although BPE does not improve the VAG-NET model for both English--German and English-French translation,
GloVe embeddings provide a substantial improvement in both language pairs.

\paragraph{Translation quality} 

To understand the model performance for translating rare words,
we computed the F-score of VAG-NET models with various debiasing techniques (Figure \ref{fscore}).
Although VAG-NET models with GloVe embeddings outperform the model with random initialization,
we do not observe a consistent improvement for rare word translation,
as reported in \cite{qi2018when}.

\paragraph{Translation examples}

Table \ref{model_examples} shows English--German translations generated by VAG-NET models with different initialization strategies.
Compared to the model without pretrained word embeddings, 
the model with GloVe embeddings generates a more fluent sentence.

\section{Conclusion}

We have explored the use of pretrained word embeddings with various multimodal NMT models.
We showed that GloVe embeddings improve the performance of all multimodal translation models,
and All-but-the-Top debiasing can result in further improvement. 

In the future, we will examine training approaches for word embeddings that are more suitable for multimodal NMT, 
especially by considering MT evaluation metrics when training word embeddings.
For example, fine-tuning word embeddings based on BLEU or other metrics for machine translation could further improve the compatibility of pretrained word embeddings with multimodal NMT models.

\section*{Acknowledgment}

This work was partially supported by JSPS Grant-in-Aid for Scientific Research (C) Grant Number JP19K12099.

\bibliographystyle{mtsummit2019}
\bibliography{mtsummit2019}

\end{document}